# Haptics in Micro- and Nano-Manipulation

**Ahmet Fatih Tabak[1] and Islam S. M. Khalil[2]**

[1]Department of Mechatronics Engineering, Istanbul Ticaret University (Turkey)

[2]Department of Biomechanical Engineering, University of Twente (The Netherlands)

*email address: aftabak AT ticaret.edu.tr, i.s.m.khalil AT utwente.nl

**Abstract**

One of the motivations for the development of wirelessly guided untethered magnetic devices (UMDs), such as microrobots and nanorobots, is the continuous demand to manipulate, sort, and assemble micro-objects with high level of accuracy and dexterity. UMDs can function as microgrippers or manipulators and move micro-objects with or without direct contact. In this case, the UMDs can be directly teleoperated by an operator using haptic tele-manipulation systems. The aim of this chapter is threefold: first, to provide a mathematical framework to design a scaled bilateral tele-manipulation system to achieve wireless actuation of micro-objects using magnetically-guided UMDs; second, to demonstrate closed-loop stability based on absolute stability theory; third, to provide experimental case studies performed on haptic devices to manipulate microrobots and assemble micro-objects. In this chapter, we are concerned with some fundamental concepts of electromagnetics and low-Reynolds number hydrodynamics to understand the stability and performance of haptic devices in micro- and nano-manipulation applications.

**Orcid IDs:**

- **Ahmet Fatih Tabak -** https://orcid.org/0000-0003-3311-6942
- **Islam S. M. Khalil -** https://orcid.org/0000-0003-0617-088X

## Introduction

Haptic tele-manipulation of microrobots and untethered magnetic devices (UMDs) is expected to have a wide spectrum of nanotechnology and nanomedicine applications. There exist many situations where it is essential to benefit from the precision of robotic systems while keeping the operator or clinician in control, as shown in Fig. 1 [1]. For example, keeping clinicians in the control loop might accelerate the adoption of this technology in clinical setting. In this case, the cognitive ability of the clinicians will be complemented by the precision and dexterity of the tele-manipulation system. Implementing such haptic tele-manipulation has proven to be a successful method for wireless manipulation of microrobots (i.e., clusters of microparticles or UMDs) [2,3]. A scaled-bilateral tele-manipulation system that allows an operator to use a haptic device to control the motion of microrobots and UMDs must enable scaling of two important physical variables. First, the motion of the operator must be scaled based on the size of the UMD, the workspace, the application intended. Second, the interaction forces between the UMD and their physical surrounding must be measured or estimated and scaled-up to be perceived by the operator. Therefore, a motion control scheme is required to allow stable tele-manipulation and successful execution of a specific task. Most tele-manipulation systems are designed to work in the macroscale world. At the macroscale, measuring the interaction forces at the remote slave-robot is less challenging than that at the microscale. Microrobots cannot be equipped with onboard force sensing capability, electronics, and transmission modules. Therefore, it is essential to design the motion control scheme of a stable tele-manipulation system based on interaction force observers.

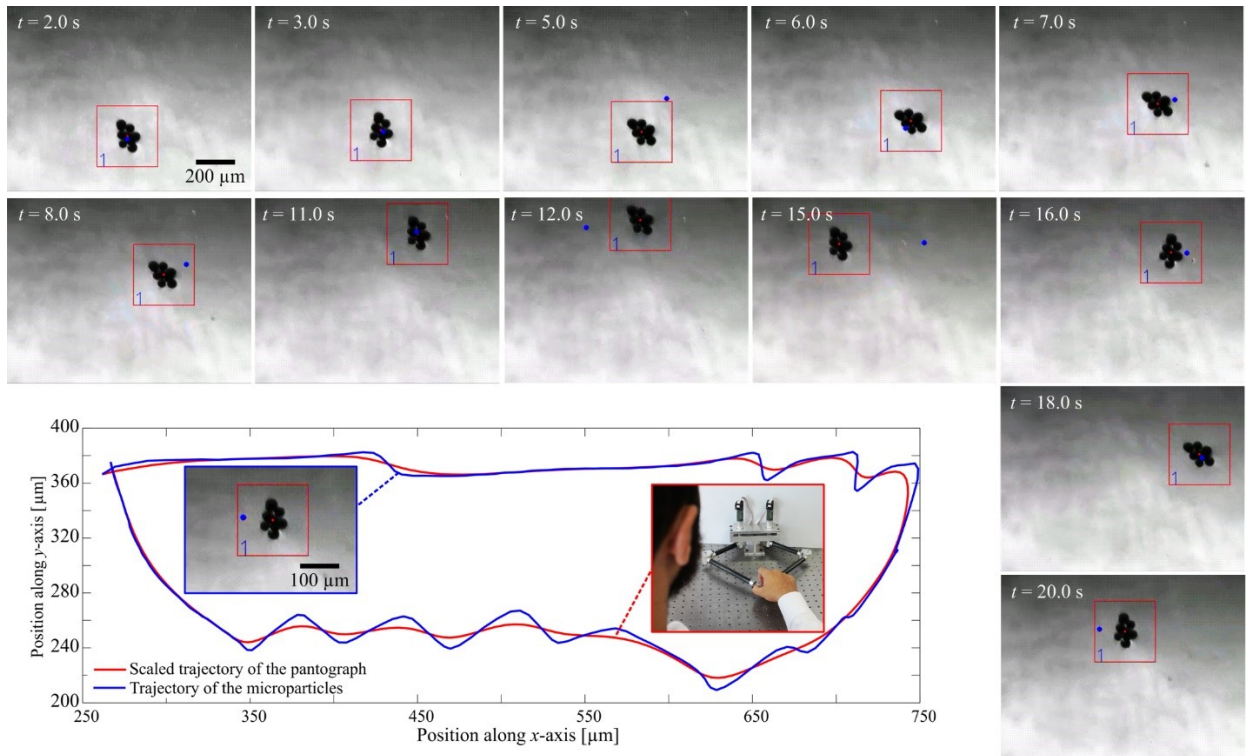

**Fig. 1.** Scaled-bilateral tele-manipulation control is achieved using a haptic device (master robot) and a cluster of paramagnetic microparticles (microrobots). The position of the end-effector is scaled down and the drag force experienced by the cluster is scaled up. The red square indicates the position of the cluster, whereas the small blue circle indicates the scaled position of the end-effector of the haptic device [3].

## Background

Closed-loop control of nano/micro-robotic devices in live tissue remains somewhat elusive and yet researchers are aiming for intracranial operations [4] or applications on various locations within the human torso [5, 6]. Although the scales are relatively smaller in comparison to most other robotic applications, problems associated with medical microrobots are not less complicated. The composite structure and impedance mismatch of stacked layers of cells offer a far more complicated engineering problem for the use of conventional sensory input [7, 8, 9, 10]. Besides, fluids of multi-phase flow conditions, filling the cavities and a most complex and interconnected network of channels, such as the circulatory system, ever in motion impede the resolution and effectiveness of detection [11, 12, 13, 14].

Medical robotics make use of teleoperation, bilateral control, and haptics to isolate the medical staff from the patient and to electromechanically couple the motion, e.g., hand gestures, with heightened robustness, precision, and repeatability to increase the effectiveness of the procedure via sensory input [15, 16]. The clinician is fed visual and tactile information while the said hand gestures are conveyed on the targeted tissue with suitably adjusted stiffness to complete given tasks [17]. The main purpose is to disturb the tissue, i.e., puncturing, cutting, or merely pushing it aside, just the right amount avoiding additional distress to the structural integrity of the surroundings. This strict condition is imposed on the design of all systems and controllers.

A robotic tool will perform as well as the controller is adequately *communicating* with the environment. However, a micro robotic tool might be devoid of the necessary volume to carry all the transducers and associated electronics [18]. The minimally invasive performance of a microrobotic tool, an untethered end-effector as opposed to kinematic chains [19], could be a definitive example of teleoperation and haptics even in the presence of medical staff along with the patient in the same room. The clinicians will have no direct tactile contact with the tissue. Furthermore, force feedback might be necessary to assure the completion of the task. However, the lack of adequate sensory input due to the absence of onboard transducers could present a unique challenge in any case. For example, a swimming microrobot in the circulatory system can be used to drill blood clots [20, 21] but shall not try to perturb the endothelial cells on the walls of the vessels physically as they respond to mechanical stress and will exhibit a reaction and could contract [22] and possibly turn the microrobot itself into a plug blocking the blood flow entirely in effect. Such a scenario would cause serious repercussions given the location, e.g., proximity to vital organs or tissue, of the microrobot at that instance [23]. Therefore, it is deemed paramount to secure *bilateral communication* to avoid such occurrences.

Teleoperation or haptics without direct electrical or mechanical coupling to transmit information without uninterrupted sensory data is a troublesome concept. Without real-time information pertaining states of the microrobot and its surroundings will lead to uncertainty of information to be processed in motion and force control loops. Hybrid solutions could be suggested to mitigate such adverse situations. Data acquisition might be achieved through a group of diverse sensors and imaging techniques [15, 24] integrated into the system and summoned ad-hoc revealing required information. In the meantime, artificial intelligence could be employed to choose the most fitting piece to complete the overall picture and reveal unique patterns associated with applications. Such an approach would be beneficial even if the microrobot is operating in a somewhat advantageous location for mere optical tracking [25]. Whenever a system state essential for the control loop simply cannot be retrieved through sensory feedback, a model-based estimator would kick in to fill the gap [26] or machine learning simply could *guess* the missing piece to this jigsaw puzzle [27] ensuring uninterrupted bilateral information flow. Therefore, a decision loop might not need to *measure* everything but it must gather as much information as possible to reduce ambiguity. However, there is yet another limitation to satisfy: the system should orchestrate the effort in real-time so that micro agents would not be lost while carrying out errands deep under composite layers of tissue. This constraint is likely to prohibit the use of cumbersome magnetic resonance imaging (MRI) systems that require time to change and modulate the magnetic field and process the information to detect spatial and temporal variations in a large volume of interest [28]. Although this restriction might not apply to a tool-tip of a catheter as it is already limited in degrees of freedom via a physical anchor and able to obtain force data for the feedback loop [29]. Certainly, when a microrobot is untethered, it can deviate from the course and disperse in the tissue if not properly observed and interfered with within a reasonable time window.

There already exist successfully practiced teleoperated-robotic applications in medicine [30]. Use of end-effectors via articulated tether, bilateral data acquisition and transmission with sensory feedback, long-distance communication, and handling data loss exceptions are widely experienced topics [31, 32, 33, 34]. Therefore, in this chapter, the focus will be on data acquisition for untethered micro robotic devices and control strategies with scarce sensory information on environmental conditions. Additionally, determining local flow conditions and material properties at possible physical contact points requires *fast-and-accurate-enough* numerical predictions. These aspects of the subject matter are also included in the discussion. Losing the track of untethered microrobots or rendering them inoperative due to the reaction of their surroundings are undesirable outcomes; therefore, as stated before, hybrid and adaptive approaches might be in demand to satisfy the requirements of a given therapeutic application.

### Tracking Microrobots: A Focused Perspective

Real-time imaging of micro robotic agents deep under layers of tissue is an engineering challenge. Therefore, keeping track of the location of microrobots and detecting their position with high resolution and certainty is at the heart of this research field [35, 36]. However, many of the motion control studies in the literature either depend on the assumption that *the necessary sensory input is somehow acquired* or the experimental study is conducted *in vitro*, i.e., an artificial container of some sort, and in *ex vivo*, i.e., samples were taken from an organism; therefore, relatively easy to visualize [37, 38]. The use of a singular type of sensory array falls short of providing adequate information for feedback loops. Although, there too exist studies with an exceptional sensory advantage such as (a) the use of x-rays for deep tissue illumination [39], which might not be feasible to operate for long periods, (b) *in vitro / ex vivo* studies under optical microscopes and cameras to obtain information for visual-feedback to achieve motion control [40, 41, 42, 43, 44, 45], and (c) studies in locations relatively much easier to observe helping in the realization of ideas at the proof-of-concept level [45, 46]. On the other hand, elaborate hybrid methods have been suggested to tackle the problem of *observing micro agents*, i.e., robots and particles, *in deep tissue*. One of the said techniques demonstrated is the 'photoacoustic computer tomography' approach merging computer tomography and detecting mechanical vibrations locally induced by absorbed light waves in soft tissue, i.e., photo-acoustics, to differentiate between various layers and materials, organic or inorganic all stacked together [47, 48, 49, 50]; therefore, allowing the tracking of microrobots. Photo-acoustics is also a method used to visualize circulatory networks in soft tissue [51]. Important takes on the experimental studies are that (a) the acoustic feedback is suitable for real-time closed-loop motion control [42, 52, 53], and (b) it is possible to actuate robotic arms and electromagnetic coils in tandem to track and control magnetic microrobots [43].

Another hybrid visualization approach magnetomotive ultrasound (MMUS) imaging method [35]. MMUS imaging is used to detect magnetically susceptible particles, i.e., reacting to time-dependent magnetic fields thus inducing mechanical vibrations that are later collected by acoustic transducers. The method offers a spatial resolution down to 10 µm in size and an advantageous penetration depth measured by a few tens of centimeters [35]. Furthermore, this technique is relatively faster than an MRI-based tracking system [35, 54]. However, the MMUS has two main limitations: (a) the temporal resolution could be close to the order of minutes, which makes it not be adequate for real-time control performance as a stand-alone system, and (b) it requires the active agents to be reacting to the magnetic field, to begin with. To put into perspective, it should be noted that the reaction speed of the photoacoustic imaging technique is much faster than MMUS but it is effective in a relatively shallow depth [35]. In a similar comparison it can also be noted that ultrasonic waves can penetrate up to 10 cm in the tissue but with a relatively lower resolution [35]. Therefore, one needs a collaboration of the systems to achieve tracking.

One of the recent *in vitro* examples in literature, although without the realization of a flexible workspace or a closed-loop control, demonstrated successful tracking of microrobots by incorporating two separate acoustic techniques [50]. The study combined ultrasonic (30 Hz) and photoacoustic (10 Hz) detection methods successfully where a single transducer and a pulse-laser were used to detect the location of active agents 25 cm deep in the bulk of a homogeneous phantom [50]. Such coordination of distinct sensory systems is demonstrated to eliminate the shortcomings, e.g., limits on resolution and penetration depth, of individual systems. However, it is important to note that the robots employed in the study are much larger than average mammalian cells that can be used as a reference dimension for microrobots [55].

The other aspect of tracking microrobots is the visualization of the surroundings for gait and route planning. Force and position control without predicting where boundaries and bifurcations are would mean reacting to circumstances as they unfold along the way. This, might not be the most intelligent and efficient way of commuting in live tissue. Thus, along with detecting the position and orientation of the microrobots, a motion control loop would also benefit from the imaging of the tissue itself. It is also possible to utilize soft tissue imaging to reconstruct the composition of the tissue in 3D [35, 48]. Furthermore, a decision loop would need the physical and geometric conditions of blood vessels to predict the best control signal.

A close look into the boundaries will affect path planning. For instance, the size of the arteries dictates that introduction of a foreign object with a comparable size would alter the flow field and pressure gradients drastically; therefore, the overall flow resistance will change [56]. In other words, the diameter of the blood veins and arteries will dictate the overall swimming performance. The vascular network can be separated into subsystems based on the size as; macrovascular, for diameter larger than 0.5 mm; mesovascular, for diameters down to 10 µm; and microvascular, for smaller diameters [57, 58]. The dimensions are important to avoid mechanical reactions [22]. The macrovascular network can be reconstructed using computer tomography (CT) scans, 3D angiography, MRI, and 3D and volumetric intravascular ultrasound (IVUS) imaging approaches [12, 57, 58, 59, 60, 61]. Also, the microvascular network, i.e., capillaries, usually surrounds the tissue as a fine mesh while the mesovascular network bifurcates that can be modeled as more of a fractal geometry [57]. The size of the blood vessels also affects the shear rates [57, 62, 63]; therefore, altering the phase of the flow locally by introducing higher viscosity and non-Newtonian effects [13]. It has been recently proposed to follow the swarm of magnetic microparticles in the bloodstream by a sweeping action of ultrasonic transducers ex vivo by Doppler imaging [52, 53]. Although the said study focuses mostly on imaging, possible adverse effects of multi-layered composite tissue between the swarm and the transducer are not addressed in detail.

At any rate, a hybrid approach may be used to visualize the tissue itself. It may be impractical to detect every cell nearby on demand to avoid attacks from the immune system [64], for instance; however, medical applications require the best possible solution. Therefore, a group of sensory systems and approaches shall be employed together to get the most complete picture possible to predict the forces on the microrobots thus yielding the most accurate force calculations. Once the position, velocity, and force data are accounted for to an acceptable degree, teleoperation, bilateral control, and haptics could be realized to meet the constraints of said medical operations.

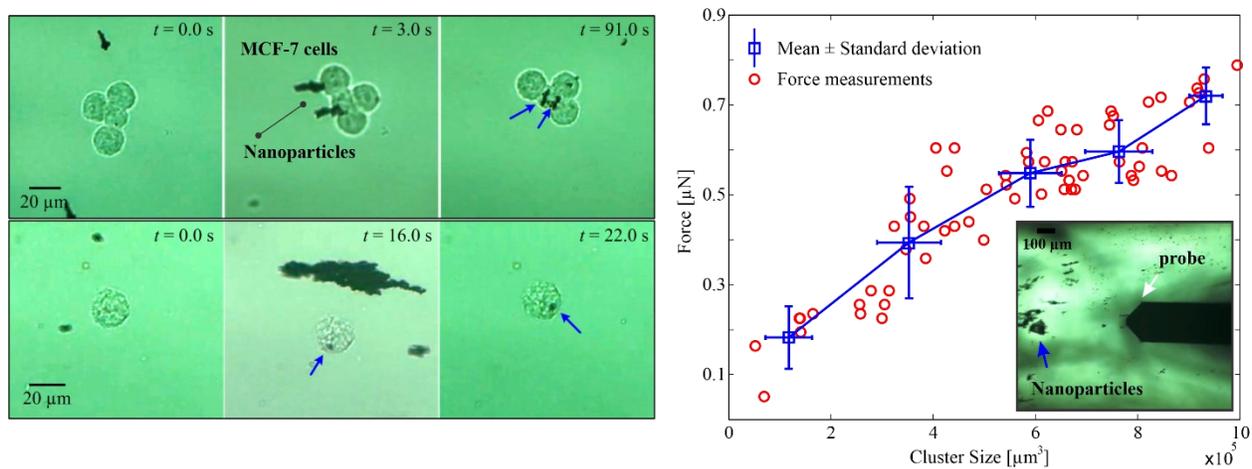

**Fig. 2.** A cluster of nanoparticles can exert force in the low micro-Newton range on cells and tissue. Left: MCF-7 breast cancer cell aggregate is targeted by cluster of iron-oxide nanoparticles using external controlled magnetic field (gradient of 5 T/m). Right: The interaction force between a microforce sensing probe and clusters of nanoparticles indicate that the force increases with the volume of the cluster for a given magnetic field gradient [65].

## Rudimentary Motion Control Scheme

Dynamics of microsystems are fundamentally different than that of their macroscale copies due to their fast reaction to stimuli. Motion control and trajectory tracking of microrobots without abundance of sensory data poses as a detrimental characteristic. For example, force measurement or estimation at microscale is challenging from an engineering perspective. Microrobots cannot be equipped with force sensing capability we cannot use the conventional sensing techniques that are used at the macroscale. These sensing techniques require onboard electronics and power. Therefore, realization of an effective control strategy, especially in the field of micro robotics of the kind discussed in this chapter, is a formidable challenge. Ideally there should exist two layers of control: (i) the automatic closed-loop control routine handling environmental effects in real-time and (ii) haptic control for the user to interfere whenever deemed necessary. For a biomedical application, haptics is the key factor to assure structural integrity of the tissue with which the robotic tool will come into contact. Haptic feedback might be assisted or preempted by the automatic closed-loop should certain conditions arise.

Arguably, biomedical micro robotic applications in arteries and other fluidics ducts are still in their infancy as we are yet to see a successful *in vivo* application. In addition to the lack of definitive sensory information there is the issue of scalability as such a system is expected to take command from medical staff and articulate the end-effector on tissue (Fig. 2). This expectation places the subject matter at the heart of *the ergonomics in robotics* for such systems should be operated *by* clinicians of diverse age, gender, physique and medical background *on patients* of diverse age, gender, physique and medical background, with maximum possible ease, comfort, and confidence. For instance, a clinician with disability should be able to practice medicine on a toddler effortlessly. Therefore, the system in question must perform reliable and feel comfortable at both ends.

We can visualize an arguably idealized control scheme as depicted in Fig 3. The user, i.e., medical staff, is delegating the task while the robot is dealing with conditions of the surrounding. The user is receiving haptic feedback from the tissue, i.e., cells, upon contact whereas the fluid environment is imposing drag regardless of the flow field. Note that, at microscale, the forces experienced by the microrobot cannot be perceived by the operator. Consider, for example, a cluster of iron-oxide nanoparticles contained with MCF-7 breast cancer cell aggregate, as shown in Fig. 2. In this case, the cluster can be considered as the end-effector of a remote tele-manipulation system in which the clinician would provide the reference trajectories and receive visual and force feedback from the environment. It is essential that force feedback is provided to the user to avoid tissue damage. Therefore, the two stimuli must be *filtered* and separated from one another. The trajectory of the microrobot can be predicted via artificial intelligence after certain number of training rounds although it is still an open question how this will be achieved to perfection without putting the tissue, and the patient, to danger. Assuming that a robotic bootcamp can be constructed and therefore the location of interest can be physically simulated elsewhere, it is theoretically possible to estimate the motion without dealing with computationally intensive calculations or hefty sensors with impossible dimensions to fit within the swimming microrobot. However, even with such a smart tool one would require (a) a method of visual tracking and (b) a mathematical model to predict and transmit the *scaled* force of the tissue on the robotic tool.

In this approach, there will be four cascaded systems. The biomedical micro robotic tool will be the smallest but the most susceptible one with direct interaction with biological fluids and tissue. Therefore, it will experience a series of generalized force components, **F**, that can be incorporated in the equation of motion as follows:

$$\mathbf{m}\ddot{\mathbf{d}} = \mathbf{F}_{drag}(\mathbf{d},\dot{\mathbf{d}}) + \mathbf{F}_{actuation}(\mathbf{d}) + \mathbf{F}_{contact}(\mathbf{d},\dot{\mathbf{d}}) + \mathbf{F}_{gravity}(\mathbf{d}), \tag{1}$$

where **m** and **d** denote the mass matrix and the vector of generalized coordinates for the microrobot. The forces experienced by the microrobot are expected to be of a function of this vector and its time rate of change. Certain fluid-drag forces, i.e., **F**$_{drag}$, will be exerted based on its relative generalized position to boundaries and relative generalized velocity to the flow field. Also, the actuation force, i.e., **F**$_{actuation}$, will be articulated by the external field applied by the intelligent control, i.e., **B** as depicted in Fig 3., as a function of its relative position. Furthermore, culmination of various electrostatic interactions with surrounding boundaries and other mobile objects in the stream is denoted by **F**$_{contact}$ as a function of relative generalized position and velocity. Finally, gravitational attraction might induce a torque component on the microrobot as denoted by **F**$_{gravity}$. Note that in low-Reynolds number fluids, viscous forces are dominant and inertial forces becomes negligible. Therefore, the dynamics of the microrobot is dominated by relatively low rates of changes that are directly dependent on the input (second term in the left-hand side of the force balance). Such input can be produced wirelessly using an external field.

The field, **B**, and its gradients can be of electromagnetic, acoustic, or optic based on the noninvasive method of choice. It is arguably more likely and practical that the generalized coordinates associated with the field will be articulated by a separate robotic positioning system (as shown in Fig. 1). The robotic positioning system can be a kinematic chain, an arrangement of electromagnetic coils, piezoelectric elements, or a light source, or a combination of these. However, there might be no tactile contact between the two systems; therefore, positioning system might possess a separate vector of generalized coordinates, i.e., **q**. Such a system can be modeled in the most general sense as follows:

$$D(q)\ddot{q} + C(q,\dot{q})\dot{q} + P\dot{q} + T\dot{q} + Kq + g(q) = u + S_{\{1,2\}}J^T(q)F(q,\dot{q}) \qquad (2)$$

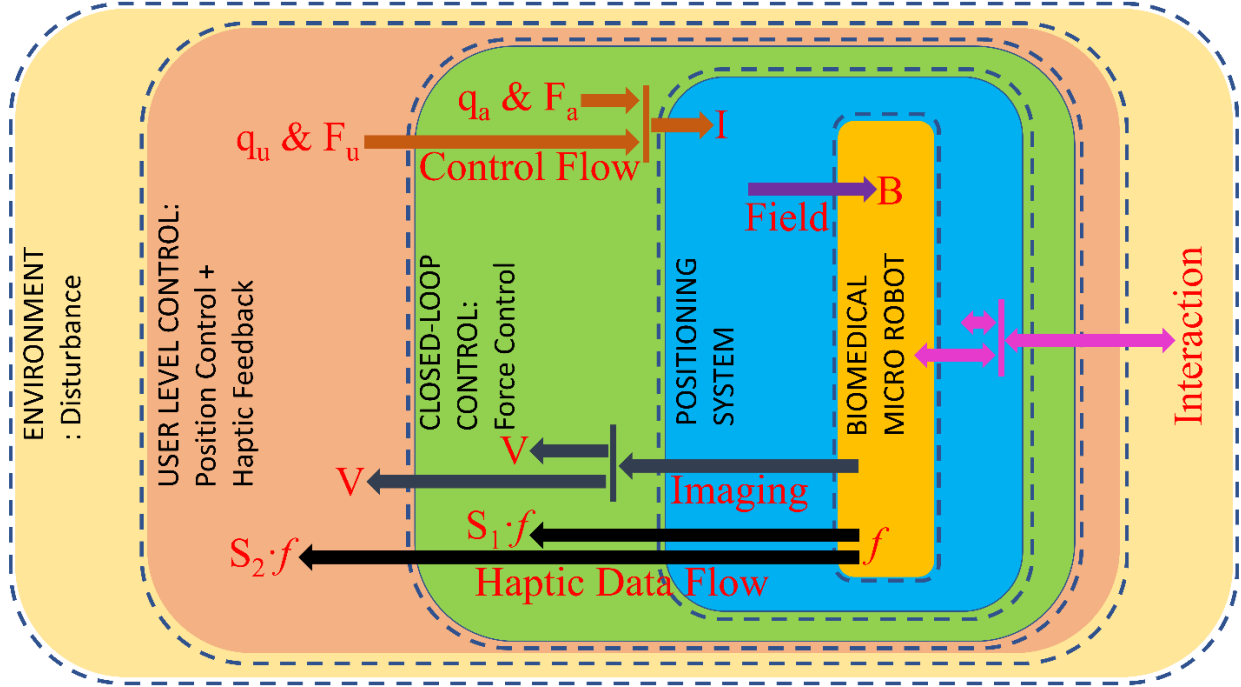

**Fig 3.** The data flow for the envisioned system with I, B, V, $S_{\{1,2\}}$, $f$, $q_a$, $F_a$, $q_u$, and $F_u$ denoting current supplied to the positioning system, the field applied on the microrobot, the visual feedback, scaling matrices for haptic feedback, the force exerted by the tissue, motion feedback from the user, force feedback from the user, motion feedback from the closed-loop controller, and the force feedback from the closed-loop controller.

with **D**, **C**, **P**, **T**, **K**, **g**, **u**, and **J** being the total inertia matrix, the matrix of the Christoffel symbols of the first kind, electromechanical friction matrix, electromechanical transducer coefficient matrix, stiffness matrix, vector of gravity, the input vector, and the manipulator Jacobian matrix. Therefore, **q** can also include electrical charge and flux linkage depending on the actuation method of the micro robotic tool [66, 67]. Here, one need to scale the force using the scale factor matrix $S_1$ if it is supposed to reflect the force not on the manipulator of the positioning system but the force on the microrobot. However, the mismatch in dimensions simply prohibits using the predicted forces in a control loop without applying scaling first so that the numbers will comparable. Scaling is an important factor as the forces experienced at the micro scale cannot be perceived by the operator. For example, the interaction forces between a cluster of nanoparticles and a microforce sensing probe are shown in Fig. 2(Right). In this experiment, clusters of iron-oxide nanoparticles controlled using field-gradient pulling toward the probe and force is measured. The measured force scales approximately linearly with the size of the cluster and remains in the low micro-Newton range. Therefore, scaling this force up is essential to allow the operator to obtain force feedback relevant to the surroundings.

The closed-loop motion control is implemented on top of the positioning system. As mentioned before, this should act as a fail-safe and prevent user to exert excess force on the surrounding tissue by simply cancelling out the user input when necessary. Therefore, it should be implemented as force control cascaded with visual-servoing. Nevertheless, the visual-servoing will not give us the position feedback;

instead, it will be used to decide to saturate the force applied by the user. As a result, the implemented controller is supposed to accept force predictions, based on Equation (1), and compare it with the maximum allowable force applied on the tissue to apply the necessary force control. The issue here is to predict the position of the micro robotic tool before it collides the tissue with enough force to damage the surface. We have shown that MFC cells can overcome penetration force more than tens of micro-Newton, as shown in Fig. 2(Left). One solution is to predict the trajectory with neural-network-powered estimators because using purely computational models might not yield fast-and-reliable-enough predictions on the fly. Only this way, arguably, it will be possible to decide whether to compensate for the user-level control or not. Such a controller can be implemented as:

$$\mathbf{u} = \mathbf{g}(\mathbf{q}) + \mathbf{S}_1 \mathbf{J}^T(\mathbf{q})\big(\mathbf{f}_d + \mathbf{k}_p \mathbf{f}_e(t) + \mathbf{k}_i \int \mathbf{f}_e(t)dt - \mathbf{k}_{damp}\dot{\mathbf{q}}\big) \tag{3}$$

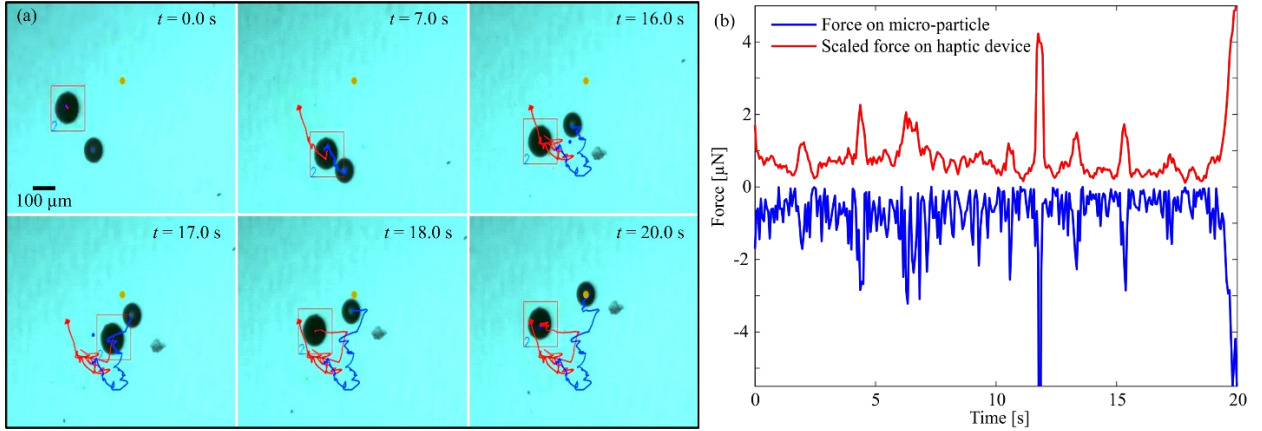

**Fig. 4.** Tele-manipulation of a nonmagnetic microbead is achieved via contact and non-contact pushing and pulling. (a) The red and blue lines indicate the paths of the slave-microrobot and the microbead, respectively. (b) The estimated interaction forces between the slave-microrobot and the microbead are scaled up and sent to the operator through the haptic device.

with $\mathbf{f}_d$, $\mathbf{f}_e$, and $\mathbf{k}_{damp}$ are the desired force vector to conduct safe operation unless it is the intention to inflict physical damage on purpose, the error in force vector, and the velocity damping to limit the acceleration on the manipulator to avoid uncontrolled velocities if the positioning system is a kinematic chain [68]. Here, the error in force is defined as $\mathbf{f}_e = \mathbf{f}_d - \mathbf{f}_{predicted}$ and it can be obtained from the solution of the very first equation. And the coefficient matrices $\mathbf{k}_p$ and $\mathbf{k}_i$ hold the necessary proportional and integral gains for PI-control loop.

The user-level control implemented on top of the closed-loop control system, depicted in Fig. 3, would make use of visual feedback to generate a control signal, thus an electrical signal, to set the position of the physical field. Such a controller architecture might be implemented as follows:

$$\mathbf{u} = \mathbf{M}(\mathbf{q})\big(\ddot{\mathbf{d}}_{desired} + \mathbf{k}_p \mathbf{d}_e(t) + \mathbf{k}_i \int \mathbf{d}_e(t)dt + \mathbf{k}_d \dot{\mathbf{d}}_e\big) + \mathbf{h}(\mathbf{q},\dot{\mathbf{q}}), \tag{4}$$

In the equation above, **M** and **h** are the *modeled* inertia matrix and the matrix containing all the rest associated with the positioning system [69] including the scaled haptic feedback from the micro robotic tool. The keyword here is *modeled* as the equation cannot represent the physical system perfectly. However, one can include as much detail with as much accuracy as possible and obtain a robust numerical approximation to the system in question. The error, i.e., $\mathbf{d}_e$, in the equation above is calculated as $\mathbf{d}_e = \mathbf{d}_{desired} - \mathbf{d}_{predicted}$ using the position of the micro robotic tool. Also, there is the derivative constant matrix, i.e., $k_d$, for the PID-control loop. And it is important to note one more time that $d_{predicted}$ is obtained not necessarily from the solution to the first equation, but imagined to be attained via neural networks based on live visual feedback.

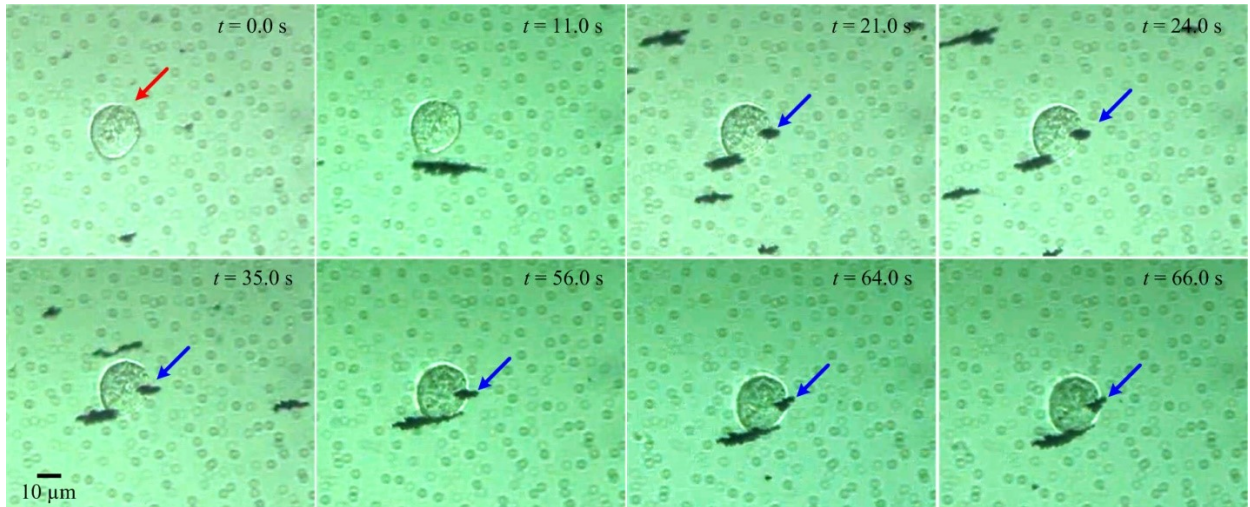

**Fig. 5**. Penetration of MCF-7 breast cancer cell (red arrow) using a cluster (blue arrow) of iron-oxide nanoparticles is achieved. The cluster is pulled towards the cell under the influence of the magnetic field gradient and penetration of the cell is achieved at time, t=21 seconds. The cluster of nanoparticles is partially engulfed by the cell at time, t=66 seconds.

Here it must be acknowledged that there will be a direct interference by the user at the user level control. We can envision that the control interface is a joystick type input device. In such a system the user, i.e., the medical staff, should experience a resistance in terms of stiffer reaction to the commands in a certain direction if the visual feedback results in excessive force predictions. That, along with the imaging, i.e., visual feedback, will give the cognitive information and tactile feeling necessary for the user. With such capability, the operator, at the macroscale world, would be able to interact with microscale objects through microrobots and achieve develop better understanding of the microscale world. For example, it would be possible to manipulate micro-objects and guide them toward unwanted cells, as we shall see later. It would also be possible to create fluidic coupling between micro-objects and implement noncontact manipulation. The first scenario is important in targeted therapeutics, where the microrobot can be loaded with a drug cargo and moved toward cell aggregate to release a concentrated dose of chemicals. The second scenario is important when interactions with a biological sample is required with minimal contamination due to direct contact.

## Usage Cases
### Micro-Manipulation of Non-Magnetic Beads

Bilateral tele-manipulation is achieved using the master-robot (inset in Fig. 1) and slave-microrobot (paramagnetic microparticle), as shown in Fig. 4. Magnetic-based tele-manipulation system is used to position non-magnetic microbeads (blue polystyrene particles, Micromod Partikeltechnologie GmbH, RostockWarnemuende, Germany). Fig. 4 provides a representative tele-manipulation experiment of the non-magnetic microbead toward a reference position (small orange circle). The positions of the slave-microrobot and the microbead are indicated using the red and blue lines, respectively (Fig. 4(a)).

Positioning of the microbead is achieved via contact and noncontact manipulation between the microbead and the slave-microrobot. At time, *t=7* seconds, the slave-microrobot touches the microbead and changes its orientation toward the reference position. At time, *t= 16* seconds, the slave-microrobot reverses its direction at a relatively high speed to break free from the adhesive force with the microbead. This action enables non-contact pushing of the microbead by moving the slave-microrobot slowly with respect the microbead toward the reference position (time instants, *t=17* seconds and *t=18* seconds). Once the microbead is positioned at the reference position (*t=20* seconds), the operator moves the slave-microrobot at relatively high speed away from the microbead to break free from the adhesive forces and to achieve a successful release. During this tele-manipulation experiment, the interaction forces are estimated, scaled-up, and sensed by the operator. The scaled-force on the master-robot and the interaction force on the slave-microrobot are shown in Fig. 4(b). In this trial, the non-contact pushing and pulling are used to accurately position the microbead within the vicinity of the reference position, and the maximum position tracking error is calculated to be 8 μm in the steady state. It is likely that this error is affected by the hydrodynamic interaction between the slave-microrobot and the nonmagnetic microbead when we attempt to move the microrobot further away from the microbead. Therefore, the positioning accuracy depends solely on the operator's inputs during each micromanipulation trial and micro assembly of several objects is challenging.

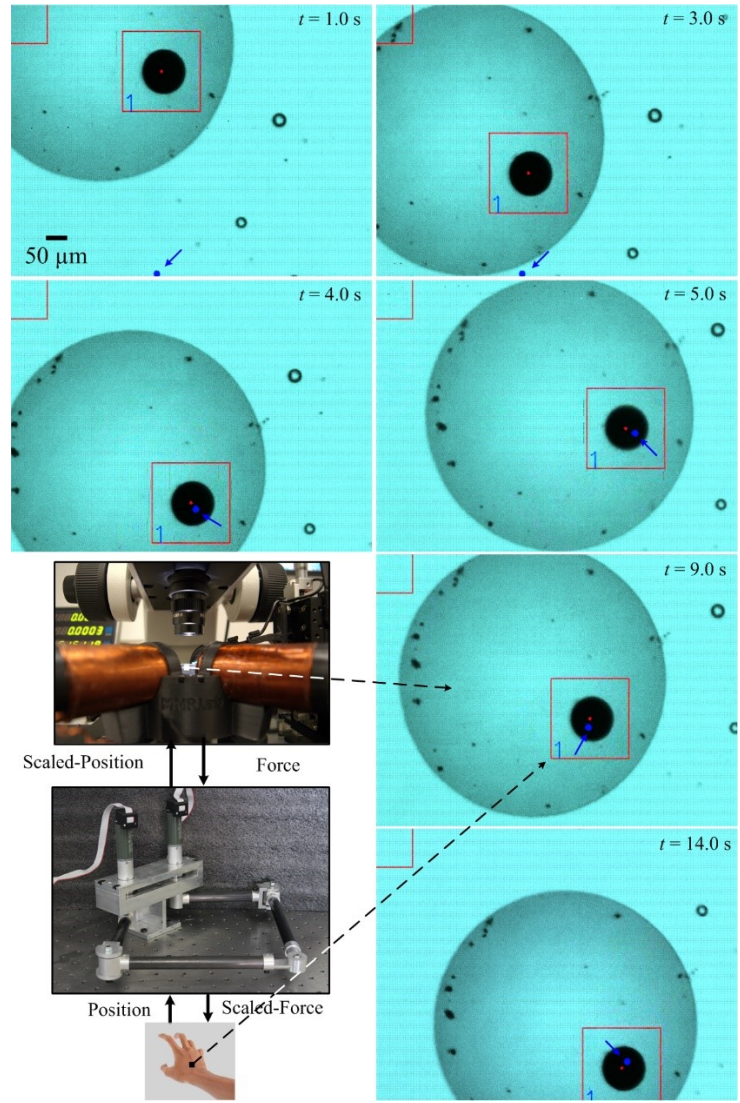

Fig. 6. A representative bilateral tele-manipulation of a gas bubble using a trapped paramagnetic micro-particle. The motion of the micro-particle is controlled under the influence of the magnetic field gradient towards the time-varying reference position (small blue circle) using a haptic device.

### Cancer Cell Penetration

In contrast to the previous use case, in which the operator used a slave-microrobot to position a nonmagnetic microbead near to a desired reference position. Targeting a cell or a specific region in tissue requires direct interaction with an environment (i.e., membrane of the cell). Penetration of the MCF-7 cells is done without incubation of the cells with the nanoparticles. The cells and clusters of nanoparticles are contained within the common center of an electromagnetic configuration (four electromagnetic coils with orthogonal configuration). The magnetic field gradient is controlled to pull the clusters toward the cells. Fig. 2 provides representative trials of the penetration of the MCF-7 breast cancer cells using clusters of nanoparticles. Two clusters with length and width of 10 µm and 4 µm are engulfed by an aggregate of cells after 64 seconds, as shown in Fig. 2 (top). The two clusters are controlled under the influence of magnetic field gradient of approximately 5 T/m. In the second trial (Fig. 2 (bottom)), single cluster with length and width of 6 µm and 4 µm, respectively, is pulled toward single MCF-7 cell and is engulfed after 22 seconds. The pre- and post-conditions of the cells indicate that the nanoparticles uptake is achieved without causing any damage to the cell membrane and without affecting the cell morphology. Fig. 5 provides another representative penetration of an MCF-7 cell using a cluster with width and length of 4 µm and 10 µm, respectively. The cell uptake of this cluster is achieved in approximately 66 seconds, and the exerted magnetic force is measured to be 0.4 µN. Achieving this level of tele-manipulation is important to test the efficacy of drugs on cancer cells. The nanoparticles could be coated with drugs, moved toward the cells to release their drug cargo. In this case, drugs will be delivered controllably to where they wanted most. Alternatively, drugs could be also contained in the polymer matrix of other types of micro or nanoparticles to achieve the same mission. It is also possible to target the cells and achieve intercellular tasks and monitor the behavior of the cells. Note also that unlike the noncontact manipulation experiment, in which the spherical slave-microrobots is used to move a nonmagnetic microbead toward a reference position. Targeting of cells is achieved with clusters of nanoparticles with relatively high aspect ratio to facilitate the uptake. Therefore, the design of the slave-microrobot is dependent on each specific application.

### Feeling Microparticles Trapped in Gas Bubbles

In the previous applications, we have demonstrated the capability of the user to implement specific micromanipulation or cell penetration missions in an intuitive manner. This method can be also used to interact with objects and increase our understanding of a physical phenomenon. Consider, for example, the interaction between a small gas bubble and a slave-microrobot. Such interaction might be useful to improve our ability to use microbubbles in medical imaging. Microbubbles are currently used to enhance ultrasonic assessment of myocardial prefusion and many other medical applications such as the ability to target cellular markers and enhance the dynamic blood flow estimation. They can also be used to deliver localized chemotherapy and potentiate some mechanisms of gene therapy, a have the potential to enhance lesion ablation through cavitation. Therefore, the interactions between a slave-microrobot and a gas microbubble is tested, as shown in Fig. 6. The haptic device allows the operator to provide reference trajectories to the controlled bubble. The electromagnetic configuration surrounds a water reservoir that contains the micro-particles and a gas bubble. The operator moves the haptic device and provides reference trajectory (indicated using the small blue circle) in 2D space. This reference trajectory is scaled-down to micro-scale to control the motion of the microparticle based on the motion of the operator. The estimated interaction force is scaled-up (by 6 orders on magnitude) to the sensory range of the operator, and sensed on the haptic device, as shown in Fig. 7.

In this application force sensing might not be as important as positioning of the microbubble. With this tele-manipulation capability, we can move microbubbles controllably in any location to improve medical imaging or diagnostics. Fig. 7 shows the estimated interaction forces between the microparticle and the gas bubble during the micromanipulation trial in Fig. 6. In this trial, it was possible to move the microbubble and cross the liquid-gas interface, allowing the microbubble to remain structurally stable for the application intended. Note that the microbubble is transparent to magnetic fields, and it is controlled via direct contact with the microparticle. It is likely that the low-Reynolds number flow contributes to the stability of these bubbles in this regime and the controlled magnetic field would allow us to selectively move microbubbles toward a desired location.

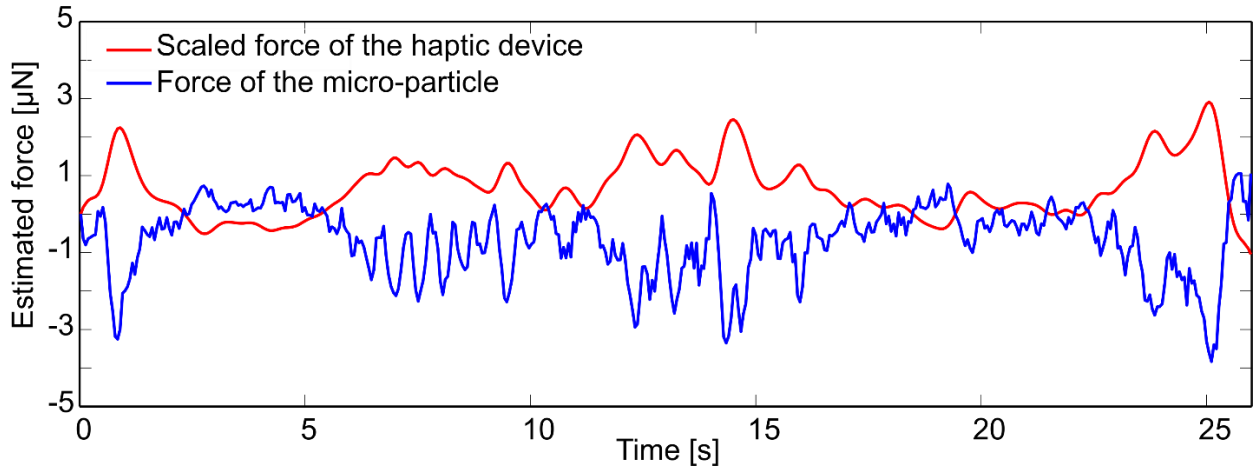

**Fig. 7.** Estimated forces of the micro-particle and the haptic device (task space forces) during a tele-manipulation trial. The red and blue lines indicate the estimated scaled force of the haptic device and the estimated force of the micro-particles, respectively.

## Conclusion

Tele-manipulation of microrobots will allow us to explore various physical phenomena and interact with the microscale world. We have shown in this chapter that a standard tele-manipulation motion control design can be used to implement a stable position control of microrobots and sense scaled bounded forces from the environment. With this level of motion control, we have demonstrated three different applications, i.e., micromanipulation of nonmagnetic objects, penetration of cancer cells, and engagement and manipulation of gas microbubbles. Developing this technology further will allow clinicians to control microrobots and reach areas in the human body inaccessible in a tethered fashion, and the impact in minimally invasive surgery and targeted therapeutic could be profound. The clinicians would be able to insert microrobots inside natural pathways inside the human body and feel their interaction with the surroundings. This capability would enable microrobots to navigate controllably toward hard-to-reach regions in the human body, while allowing clinicians with control and sensing capabilities simultaneously.